\documentclass[english]{article}
\usepackage[T1]{fontenc}
\usepackage[latin9]{inputenc}
\usepackage{geometry}
\usepackage{babel}
\usepackage{booktabs}
\usepackage{amstext}
\usepackage{amssymb}
\usepackage{graphicx}
\usepackage[unicode=true,pdfusetitle,
 bookmarks=true,bookmarksnumbered=false,bookmarksopen=false,
 breaklinks=false,pdfborder={0 0 1},backref=false,colorlinks=false]
 {hyperref}

\makeatletter

\providecommand{\tabularnewline}{\\}

\renewcommand{\vec}[1]{\mathbf{#1}}

\providecommand{\keywords}[1]
{
  \small	
  \textbf{\textit{Keywords---}} #1
}

\makeatother

\begin{document}
\title{Binary Stochastic Filtering: feature selection and beyond}
\author{Andrii Trelin \\
\texttt{trelinn@vscht.cz}\and Aleš Procházka \\
\texttt{a.prochazka@ieee.org}}
\maketitle
\begin{abstract}
Feature selection is one of the most decisive tools in understanding
data and machine learning models. Among other methods, sparsity induced
by $L^{1}$ penalty is one of the simplest and best studied approaches
to this problem. Although such regularization is frequently used in
neural networks to achieve sparsity of weights or unit activations,
it is unclear how it can be employed in the feature selection problem.
This work aims at introducing the ability to automatically select
features into neural networks by rethinking how the sparsity regularization
can be used, namely, by stochastically penalizing feature involvement
instead of the layer weights. The proposed method has demonstrated
superior efficiency when compared to a few classical methods, achieved
with minimal or no computational overhead, and can be directly applied
to any existing architecture. Besides, the method is easily generalizable
for neuron pruning and selection of regions of importance for spectral
data.

\keywords{neural network, feature selection, neuron pruning}
\end{abstract}

\section{Introduction}

Feature selection is of great interest in all machine learning task,
since it reduces the computational complexity of the models, frequently
improves generalization, and helps in data understanding. In general,
feature selection methods are divided into the following categories
\cite{chandrashekar2014survey}:
\begin{description}
\item [{Filter~methods}] use feature metrics, such as correlation, information
gain to distinguish between useful and useless features. 
\item [{Wrapper~methods}] use the feedback of model metrics to optimize
the selected feature subset. This problem can be exactly solved only
by brute force, which makes it intractable in the majority of cases.
Numerous heuristics are suggested (modern researches mainly focused
on swarm intelligence optimization \cite{gu2018feature,hancer2018pareto,mafarja2018evolutionary}),
but they are can not guarantee optimality.
\item [{Embedded~methods}] exists for certain algorithms that create a
feature importance score during training. Classical examples are decision
tree-based algorithms and $L^{1}$-penalized linear models.
\end{description}
It is obvious that models able to automatically find optimal features
are the most desired type of feature selector since it provides both
trained model and important features subset simultaneously. Unfortunately,
that is usually possible only for very simple models, while deep neural
networks (NN), one of the most crucial state-of-the-art algorithms,
are unable to perform feature selection during training. The presented
paper is devoted to the development of method to resolve that issue
by augmenting the network with stochastic variant of $L^{1}$ penalization,
which can be interpreted as stochastic search in the feature space.

\section{$L^{1}$ penalization for neural networks}

The most straightforward way of how to achieve sparsity with neural
networks is to add $L^{1}$ penalty . This method is widely used to
achieve representation sparsity \cite{glorot2011deep,ng2011sparse}
by penalizing neuron activations or sparsity of convolutional kernels
\cite{liu2015sparse,engelcke2017vote3deep} that improves performance
of convolutional models. Although $L^{1}$ penalization efficiently
sparsifies networks, the structure of the obtained sparse representation
is unpredictable and thus can not be used for feature selection or
neuron pruning. The work of Wen et al.\cite{wen2016learning} handles
that issue by explicitly introducing structure, penalizing individual
components of network such as channels, layers, etc. At the same time,
$L^{1}$ penalization for feature selection has not been yet applied
to neural networks.

We suggest how the well-known sparsity constraints can be applied
to neural networks input aiming feature selection. The proposed method
exhibits high universality and can be applied to selection of input
features, convolutional kernels, regions of importance, etc. It should
not be confused with widely used weights or activation regularization.

\section{Related works}

Sparsification of neural networks is a popular research subject of
significant importance, since it allows to make large and computationally
demanding neural networks smaller and more efficient to be run on
mobile devices. Application of structured $L^{1}$ penalty for optimization
of neural network architecture was suggested by Wen et al. \cite{wen2016learning}
and Scardapane et al. \cite{scardapane2017group}. Both approaches
are deterministic.

Since the proposed method is stochastic, it shares common properties
with a wide variety of stochastic regularization technics, derived
from the original Dropout \cite{srivastava2014dropout}. Energy-based
dropout \cite{salehinejad2020edropout} regularizes and prunes network
by optimizing scalar \textit{energy} with differential evolution algorithm.
Work of Srinivas et al. \cite{srinivas2016generalized}defines a family
of Dropout-like techniques. One of them, Dropout++ uses stochastic
neuron dropping with trainable parameters, derived through Bayesian
NN, that lead to similar although not identical formulation of filtering
units. Adaptive Dropout \cite{ba2013adaptive} achieves tuning of
dropping probabilities by augmenting neural network with binary belief
network.

\section{Binary stochastic filtering}

The main idea of the proposed method (BSF) is application of $L^{1}$
penalty on the involvement of the variable in the training/prediction
process. This is done by element-wise multiplying of input datum $\vec{x}$
by the random vector $\vec{r}$ such that $r_{i}\sim Bernoulli(p_{i})$,
where vector $\vec{p}$ defines a tunable set of parameters. This
is similar to the Dropout technic, which performs the same multiplication,
but its weights are predefined constant. Vector $\vec{p}$ is penalized
with $L^{1}$ norm, which stochastically forces the model to use only
the most important features. Another way to imagine it is to think
about it as stochastic investigation of parameter space, which at
the same time penalizes the number of involved features.
\global\long\def\expect{\mathbb{E}}%

\paragraph{Gradients}

To make the layer weights $\vec{{p}}$ trainable, it is necessary
to define two gradients for backpropagation to work, namely, $\nabla_{x}\text{bsf}(\vec{{x}})$
and $\nabla_{p}\text{bsf}(\vec{x})$, where $\text{bsf}(\vec{x})=\vec{x}\circ\vec{r}$.
We define the first gradient as 
\[
\frac{\partial\text{bsf}(x_{i})}{\partial x_{i}}=\frac{\partial x_{i}r_{i}}{\partial x_{i}}=r_{i},
\]
which is a natural way to describe a variable passed or dropped, similarly
to the Dropout. It is more tricky is to define $\nabla_{p}\text{bsf}(\vec{x})$
due to its randomness. Instead, we can differentiate the expected
value
\[
\frac{\partial\expect\text{bsf}(x_{i})}{\partial p_{i}}=\frac{\partial\expect x_{i}r_{i}}{\partial p_{i}}=\frac{\partial x_{i}\expect r_{i}}{\partial p_{i}}=\frac{\partial x_{i}p_{i}}{\partial p_{i}}=x_{i}
\]
and use it as gradient estimate. Moreover, it was empirically found
that it is useful to scale the gradient by the weight value, i.e.
to redefine the gradient as $\frac{\partial\text{bsf}(x_{i})}{\partial p_{i}}=x_{i}p_{i}$.
This modification has a clear interpretation: the lower weight $p_{i}$
the lower is feature involvement in the training process, thus the
weights of this feature must be changed slower. This modification
stabilizes training and prevents already disabled features from being
re-enabled. 

A behavior of the filtering layer during inference phase is altered
by setting a threshold $\tau$ and deterministically passing features
above threshold, while features corresponding to weights below threshold
are dropped. This replacement makes the layer at inference phase deterministic,
which stabilizes validation metrics. Implementation of BSF layer in
TensorFlow 2 framework can be found in the repository\footnote{\url{https://github.com/Trel725/BSFilter}}.

\paragraph{Analysis}

To get some understanding of how this method work we will investigate
its behavior on the simplest possible model -- linear regression.
We will start with the general formula for linear regression 
\[
\min_{\vec{w}}||\vec{y}-\mathbb{X}\vec{w}||^{2},
\]
where $\vec{y}$ is a vector of target values, $\vec{w}$ is a vector
of model weights, and $\mathbb{X}$ is a matrix of input data, such
that each row of the matrix $\mathbb{X}_{\cdot i}$ is a single observation
vector $\vec{x}_{i}$. Now, our goal is to investigate how will the
optimization objective change if we multiply each $\vec{x}_{i}$ by
a random vector $\vec{r}$. Since our objective is now random, we
will minimize its expected value, i.e. 
\[
\min_{\vec{w}}\expect||\vec{y}-(\mathbb{R}\circ\mathbb{X})\vec{w}||^{2},
\]
where $\mathbb{R}$ is a matrix, such that $\mathbb{R}_{ij}\sim Bernoulli(p_{j})$.
It can be shown (the derivation of the equation below is given in
the supporting information) that the optimization objective is equivalent
to 
\[
\min_{\vec{w}}||\vec{y}-\mathbb{X}(\vec{w}\circ\vec{p})||^{2}+||\Gamma(\vec{w}\circ\sqrt{\vec{p}\circ(1-\vec{p})})||^{2}+\lambda||\vec{p}||,
\]
 where $\Gamma=\text{diag}(\mathbb{X}^{T}\mathbb{X})^{1/2}$ , i.e.
its diagonal elements correspond to standard deviations of features
in $\mathbb{X}$ (supposing they are centered), $\circ$ denotes Hadamard
product. We can see that if $p_{i}=p$, the $p(1-p)$ member can be
taken out of the norm expression, which gives an identical expression
to the one derived in \cite{srivastava2014dropout} (when $\lambda=0$).
From that objective we can get some insights of model behavior:
\begin{enumerate}
\item For $p_{i}\in(0,1)$ the $i$th feature is efficiently penalized with
$L^{2}$ norm, where the penalty is additionally scaled by the standard
deviation of that feature. Thus, weights for the strongly varying
feature are penalized more, which is similar to classical Dropout.
\item If $p_{i}=0$, which is forced by the $L^{1}$ penalty, or $p_{i}=1$,
the middle term vanishes and weight of $i$th feature is not penalized.
\end{enumerate}

\paragraph{Stochastic vs deterministic}

It is not immediately clear why to prefer stochastic regularization
to deterministic. Firstly, weights penalization is clearly enough
to achieve sparsity for simple shallow models like Lasso regression.
At the same time, deep models can efficiently rescale back near zero
features in the hidden layers. Stochastic regularization is free from
that issue since it has only two possible states, feature is either
passed without changes or set to zero. Moreover, it is well known
in the machine learning literature that addition of noise to the network
has positive effects on model generalization and convergence \cite{rifai2011adding,bengio2013estimating,neelakantan2015adding}.
It was observed in experiments that stochastic models are actually
more stable at training phase and produce better separated (into important
and unimportant) features. An example of the model convergence curves
and selected feature importances is given in the Fig. \ref{fig:Change-in-F1},
left. 

\section{Experiments}

Binary stochastic filtering layer was implemented in TensorFlow 2
framework \cite{abadi2016tensorflow} according to the definition
above. A collection of datasets from OpenML-CC18 benchmark suite \cite{bischl2017openml}
was used in the experiments. It contains 72 classification datasets
that satisfy a number of desired properties, including balancing,
reasonable number of features and observations, moderate classification
difficulty, etc. Moreover, the authors provided reference preprocessing
and cross-validation splitting, which facilitates replication of experiments.
NN models typically require tuning of hyperparameters to get fair
results, thus a subset of 10 datasets was selected from the OpenML-CC18
and used in further experiments. Threshold $\tau$ was set to 0.25
and $\text{F}_{1}$ score was used as the main evaluation metric in
all experiments. 

\subsection{Feature selection}

\begin{figure}
\begin{centering}
\includegraphics[width=2.8in]{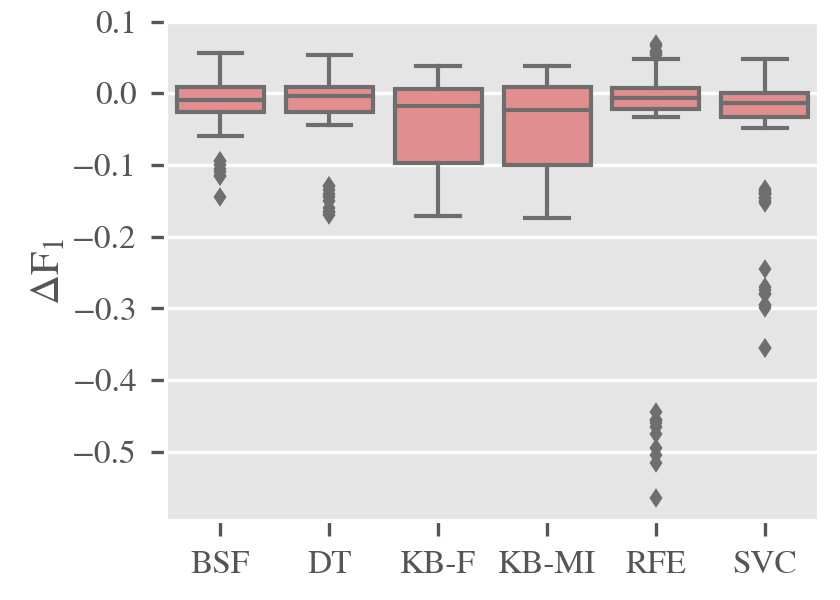}\quad{}\includegraphics[width=2.8in]{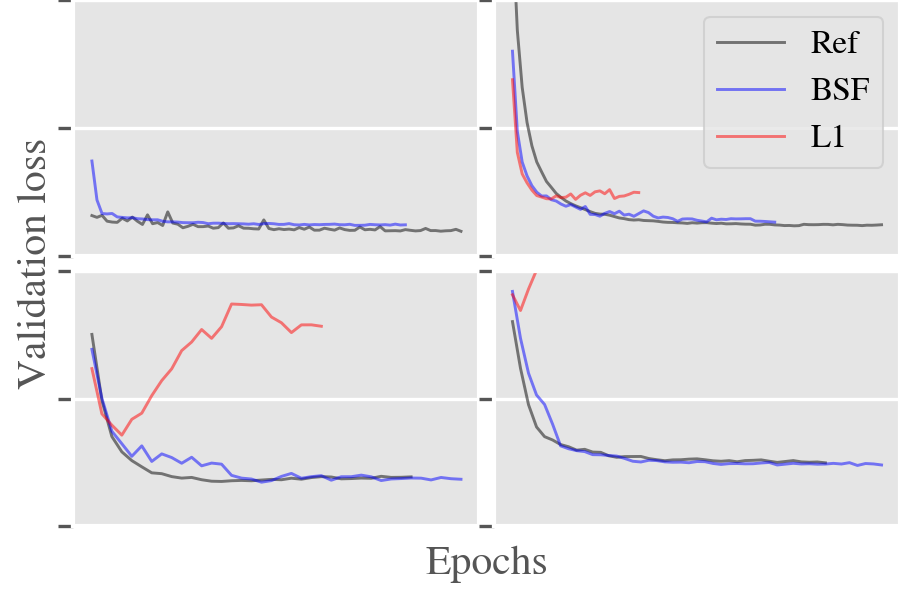}
\par\end{centering}
\caption{\label{fig:Change-in-F1}Change in $\text{F}_{1}$ score after feature
selection with different methods, sorted in ascending order according
to the group means (left). Examples of validation loss evolution for
reference model and $L^{1}$ penalized models. Early stopping after
20 epochs without loss improvement was used (right).}

\end{figure}
\begin{table}
\begin{centering}
\begin{tabular}{cccccccc}
\toprule 
ID & BSF & DT & KB-F & KB-MI & RFE & SVC & Features\tabularnewline
\midrule
\midrule 
16 & -0.0995 & -0.1490 & -0.1285 & -0.1280 & -0.4835 & -0.2950 & 6/64\tabularnewline
\midrule 
32 & -0.0014 & 0.0007 & -0.0058 & -0.0063 & -0.0019 & -0.0055 & 13/16\tabularnewline
\midrule 
45 & 0.0169 & 0.0169 & 0.0191 & 0.0185 & -0.0031 & -0.0053 & 6/60\tabularnewline
\midrule 
219 & 0.0371 & 0.0271 & 0.0217 & 0.0230 & 0.0218 & 0.0375 & 7/8\tabularnewline
\midrule 
3481 & -0.0213 & -0.0303 & -0.1445 & -0.1468 & -0.0182 & -0.0355 & 56/617\tabularnewline
\midrule 
9910 & 0.0192 & 0.0080 & 0.0056 & 0.0061 & 0.0357 & 0.0075 & 166/1776\tabularnewline
\midrule 
9957 & 0.0057 & 0.0048 & 0.0009 & -0.0010 & 0.0114 & 0.0048 & 23/41\tabularnewline
\midrule 
9977 & -0.0333 & -0.0187 & -0.0606 & -0.0607 & -0.0218 & -0.0180 & 7/118\tabularnewline
\midrule 
14952 & -0.0131 & -0.0024 & -0.0116 & -0.0111 & -0.0194 & -0.0149 & 15/30\tabularnewline
\midrule 
146825 & -0.0244 & -0.0304 & -0.1025 & -0.1027 & --- & -0.1425 & 102/784\tabularnewline
\midrule 
167140 & -0.0053 & -0.0050 & -0.0822 & -0.0813 & -0.0031 & -0.0057 & 10/180\tabularnewline
\bottomrule
\end{tabular}\caption{\label{tab:Mean-differences}Mean differences between metrics for
model trained on full and feature-selected datasets.}
\par\end{centering}
\end{table}

For the main experiment features were selected from each experimental
dataset by training a penalized model. The penalization coefficient
was manually tuned to achieve maximal reduce in number of features,
while keeping metrics reasonable. Other popular methods (implemented
in scikit-learn \cite{scikit-learn}) were selected for comparison,
corresponding abbreviations are given in parentheses:
\begin{itemize}
\item Filtering features based on mutual information (KB-MI) and ANOVA F-value
(KB-F)
\item Recursive feature elimination with SVM as a base classifier (RFE)
\cite{guyon2002gene}
\item Embedded methods: $L^{1}$ penalized SVM (SVC) and decision tree (CART
algorithm, DT)
\end{itemize}
The same number of features was selected with these methods and NN
model was trained on each of the selected feature subsets. Metrics
for each cross-validation split were collected and differences between
reference full-featured score and feature-selected one were used as
a measure of feature selection efficiency. Cross-validation splits
were same for all experiments. Results are provided in Fig. \ref{fig:Change-in-F1},
which visualizes the distribution of $\Delta\text{F}_{1}=\text{F}_{1,\text{fs}}-\text{F}_{1,\text{ref}}$,
i.e. positive values correspond to feature selected $\text{F}_{1}$
score higher than original one. It follows from the data that BSF
leads to the lowest decrease of classification score. Although the
difference with its closest rival (DT) is small, it is statistically
significant with Wilcoxon test p-value $=0.0233$. Exact values are
tabulated in the Tab. \ref{tab:Mean-differences}\footnote{RFE feature selection for dataset 146825 was intractable, thus this
value is missing from the table}. It is important to note that augmenting model with BSF layer has
only minor impact of its convergence (Fig. \ref{fig:Change-in-F1},
left), thus the filtering layer can be added to any model almost without
overhead

\subsection{Neuron pruning}

\begin{figure}[p]
\begin{centering}
\includegraphics[width=5in]{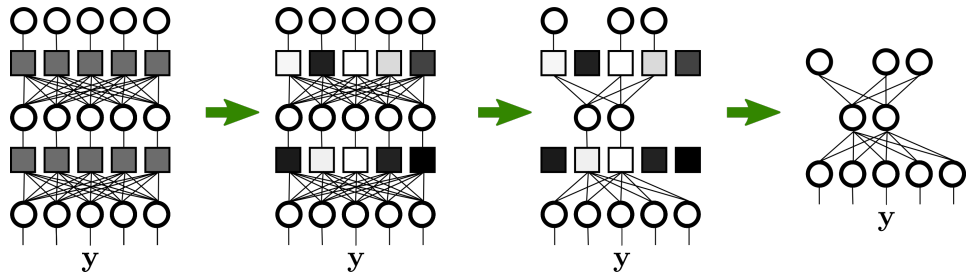}
\par\end{centering}
\caption{\label{fig:Visualization-of-pruning}Visualization of pruning with
BSF. Neurons and BSF units are drawn in circles and squares respectively.
Weights of BSF are shown as saturation of squares fill.}

\end{figure}
\begin{figure}[p]
\begin{centering}
\includegraphics[width=2.8in]{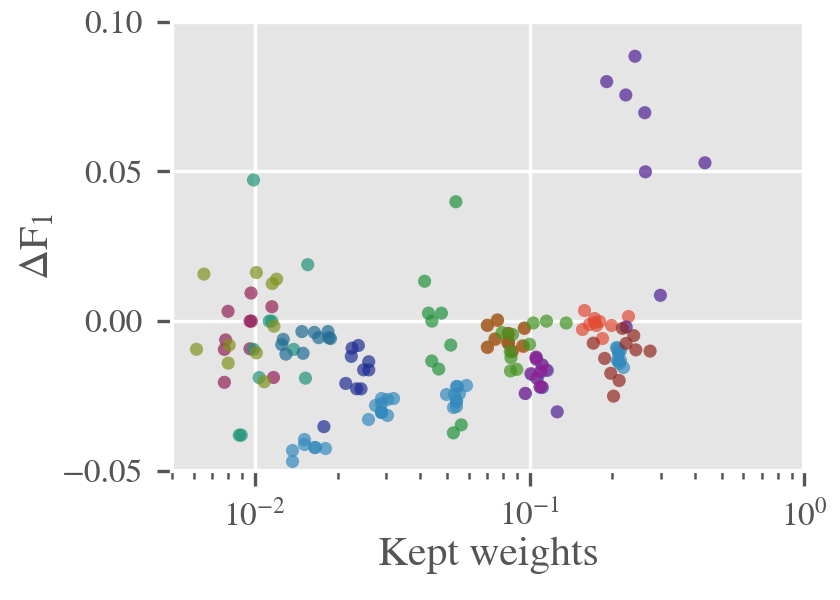}\quad{}\includegraphics[width=2.8in]{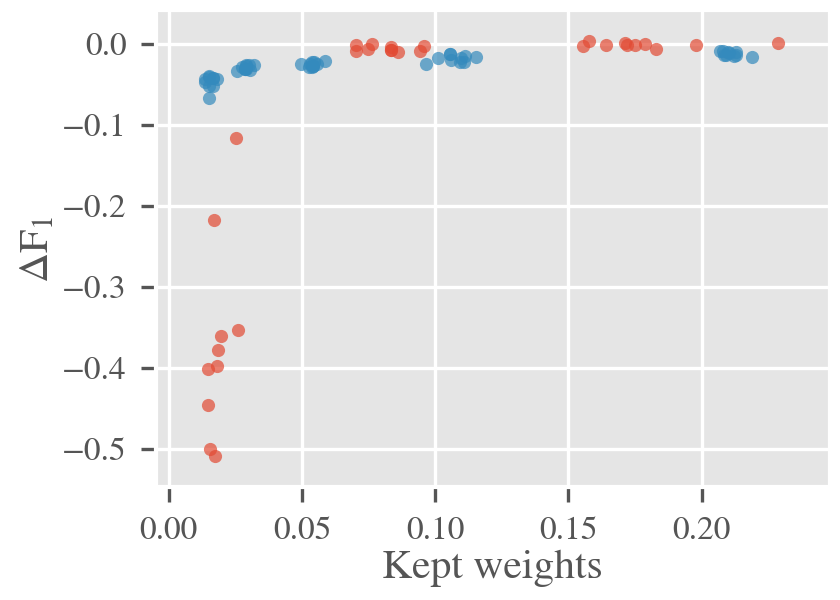}
\par\end{centering}
\caption{\label{fig:F1_pruning}Change in classification metrics after pruning
vs fraction of kept weights. Values for optimized regularization coefficient
for all datasets (left); trade-off between model accuracy and complexity
for different regularization coefficients for two selected datsets
(right). Datset IDs are represented in colors.}
\end{figure}
For the second experiment every dropout layer was replaced with penalized
BSF layer. Regularization coefficient was shared among layers, but
normalized by the starting number of neurons in the layer to achieve
equal penalization. Every model was trained on the same selected datasets,
the BSF layers were removed, and neurons, corresponding to the low
BSF values were pruned, which was achieved by removing corresponding
columns and/or rows from the weight matrix for each layer (Fig. \ref{fig:Visualization-of-pruning}).
Differences in $\text{F}_{1}$ score for the obtained pruned model
are plotted against the relative amount of kept weights in Fig. \ref{fig:F1_pruning}.
The same figure demonstrates how the number of weights can be further
decreased by the price of reduce in classification metrics.

\subsection{Region selection in spectra}

\begin{figure}
\includegraphics[width=2.8in]{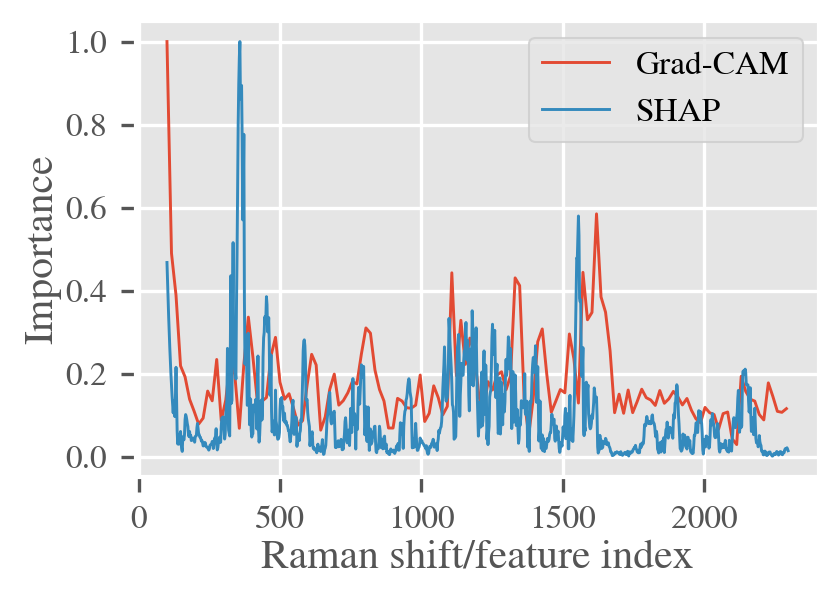}\quad{}\includegraphics[width=2.8in]{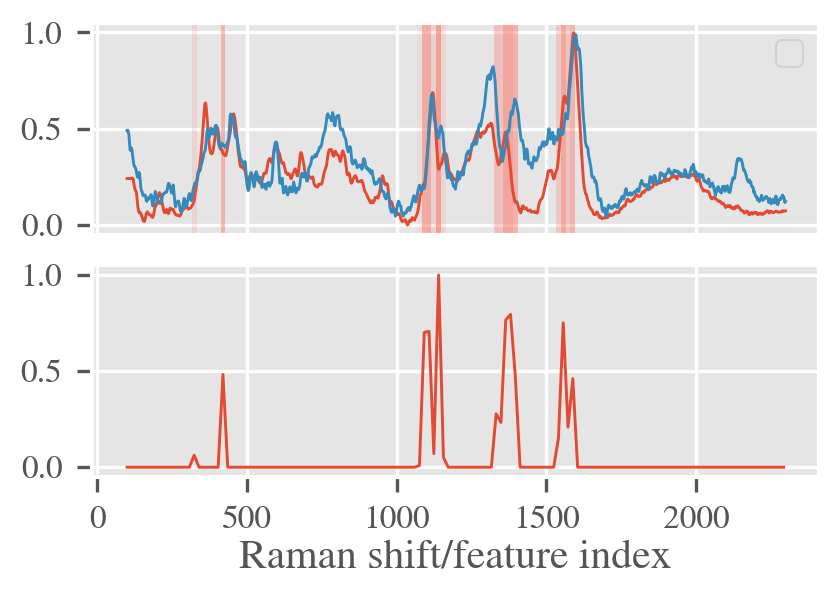}

\caption{\label{fig:Selected-regions-of-importance}Selected regions of importance
with Grad-CAM and SHAP methods (left); Regions of importance selected
by BSF. For visualization, two Raman spectra (of human glycoproteins)
from dataset are plotted above with selected regions highlighted in
red.}

\end{figure}
Spectra are one of the most common data in natural sciences. Automated
recognition of spectra is highly usesul in all branches of chemistry
\cite{ghosh2019deep,cui2018modern,zhang2019deepspectra} and biology
or medicine \cite{sigurdsson2004detection,chen2004artificial,charvat2020diffuse}.
Such signals share important property, existence of importance regions,
areas which are crucial for their interpretation. While for images
relative positions of features matter (which are usually extracted
with convolutional layers), spectra are recognized based on global
positions of peaks or other features. Although it may seem like a
problem for which fully-connected network is more suitable, convolutional
layers are still advantageous for processing spectral information
since they learn preprocessing of data, such as background subtraction,
noise filtering, etc. Extraction of the importance regions from spectral
data is exceptionally useful, since it sheds light on the processes
that generate the data. Numerous approaches were proposed to highlight
most salient regions aiming explanation of model decisions, including
Grad-CAM \cite{selvaraju2017grad}, LIME \cite{lime} and SHAP \cite{lundberg2017unified}.
Unfortunately, these methods, developed to explain individual predictions,
frequently produce overly complicated picture, highlighting noise
and clearly useless regions. Combination of individual explanation
to get dataset-wise explanation is nontrivial and its interpretation
is frequently unclear. 

Although this problem can be formulated as classical feature selection,
it is a poor approach since it disrupts the continuity of the spectra
and breaks the convolutional preprocessing. Desired importance regions
selection can be accomplished by selecting features at the output
of convolutional counterpart of network, which can be performed with
BSF layer that shares weights along the channels axis. For experiment,
the custom Raman spectra dataset of glycoproteins was classified with
simple convolutional classifier, and obtained importance regions were
analyzed with Grad-CAM, adapted for analysis of 1D convolutional networks,
SHAP explainer and BSF. The obtained results are presented in the
Fig. \ref{fig:Selected-regions-of-importance}. As it was mentioned
above, SHAP and Grad-CAM detections of region importances are cumbersome
and practically useless, while BSF has clearly selected the most informative
regions which has clear chemical interpretation. This approach was
successfully used in two analytical projects \cite{guselnikova2019label,erzina2020precise}.

\section{Conclusion}

The conducted experiments demonstrated that BSF selects features at
least as efficiently as best of the classical methods. At the same
time, it can be embedded directly in the NN model, eliminating the
need for external feature selector. Moreover, thanks to its differentiability
it can be utilized not only to drop nodes from the input layer (i.e.
features) but can be placed in the middle of the model, which can
be utilized for neuron pruning. This approach is also applicable for
filtering of convolutional channels by simple weight sharing of the
BSF layer along all axes except channel axis. Instead, if selection
of regions of importance is an aim, BSF can be applied by sharing
weights along channels axis. It was shown that for some datasets this
method allows to reduce network size to approximately 1\% of the original
size without significant reduce of classification metrics. BSF has
potential to become an indispensable tool for processing of spectral
data, particularly valuable in natural sciences.


\begin{thebibliography}{10}

\bibitem{chandrashekar2014survey}
Girish Chandrashekar and Ferat Sahin.
\newblock A survey on feature selection methods.
\newblock {\em Computers \& Electrical Engineering}, 40(1):16--28, 2014.

\bibitem{gu2018feature}
Shenkai Gu, Ran Cheng, and Yaochu Jin.
\newblock Feature selection for high-dimensional classification using a
  competitive swarm optimizer.
\newblock {\em Soft Computing}, 22(3):811--822, 2018.

\bibitem{hancer2018pareto}
Emrah Hancer, Bing Xue, Mengjie Zhang, Dervis Karaboga, and Bahriye Akay.
\newblock Pareto front feature selection based on artificial bee colony
  optimization.
\newblock {\em Information Sciences}, 422:462--479, 2018.

\bibitem{mafarja2018evolutionary}
Majdi Mafarja, Ibrahim Aljarah, Ali~Asghar Heidari, Abdelaziz~I Hammouri,
  Hossam Faris, Al-Zoubi Ala'M, and Seyedali Mirjalili.
\newblock Evolutionary population dynamics and grasshopper optimization
  approaches for feature selection problems.
\newblock {\em Knowledge-Based Systems}, 145:25--45, 2018.

\bibitem{glorot2011deep}
Xavier Glorot, Antoine Bordes, and Yoshua Bengio.
\newblock Deep sparse rectifier neural networks.
\newblock In {\em Proceedings of the fourteenth international conference on
  artificial intelligence and statistics}, pages 315--323, 2011.

\bibitem{ng2011sparse}
Andrew Ng et~al.
\newblock Sparse autoencoder.
\newblock {\em CS294A Lecture notes}, 72(2011):1--19, 2011.

\bibitem{liu2015sparse}
Baoyuan Liu, Min Wang, Hassan Foroosh, Marshall Tappen, and Marianna Pensky.
\newblock Sparse convolutional neural networks.
\newblock In {\em Proceedings of the IEEE conference on computer vision and
  pattern recognition}, pages 806--814, 2015.

\bibitem{engelcke2017vote3deep}
Martin Engelcke, Dushyant Rao, Dominic~Zeng Wang, Chi~Hay Tong, and Ingmar
  Posner.
\newblock Vote3deep: Fast object detection in 3d point clouds using efficient
  convolutional neural networks.
\newblock In {\em 2017 IEEE International Conference on Robotics and Automation
  (ICRA)}, pages 1355--1361. IEEE, 2017.

\bibitem{wen2016learning}
Wei Wen, Chunpeng Wu, Yandan Wang, Yiran Chen, and Hai Li.
\newblock Learning structured sparsity in deep neural networks.
\newblock In {\em Advances in neural information processing systems}, pages
  2074--2082, 2016.

\bibitem{scardapane2017group}
Simone Scardapane, Danilo Comminiello, Amir Hussain, and Aurelio Uncini.
\newblock Group sparse regularization for deep neural networks.
\newblock {\em Neurocomputing}, 241:81--89, 2017.

\bibitem{srivastava2014dropout}
Nitish Srivastava, Geoffrey Hinton, Alex Krizhevsky, Ilya Sutskever, and Ruslan
  Salakhutdinov.
\newblock Dropout: a simple way to prevent neural networks from overfitting.
\newblock {\em The journal of machine learning research}, 15(1):1929--1958,
  2014.

\bibitem{salehinejad2020edropout}
Hojjat Salehinejad and Shahrokh Valaee.
\newblock Edropout: Energy-based dropout and pruning of deep neural networks.
\newblock {\em arXiv preprint arXiv:2006.04270}, 2020.

\bibitem{srinivas2016generalized}
Suraj Srinivas and R~Venkatesh Babu.
\newblock Generalized dropout.
\newblock {\em arXiv preprint arXiv:1611.06791}, 2016.

\bibitem{ba2013adaptive}
Jimmy Ba and Brendan Frey.
\newblock Adaptive dropout for training deep neural networks.
\newblock In {\em Advances in neural information processing systems}, pages
  3084--3092, 2013.

\bibitem{rifai2011adding}
Salah Rifai, Xavier Glorot, Yoshua Bengio, and Pascal Vincent.
\newblock Adding noise to the input of a model trained with a regularized
  objective.
\newblock {\em arXiv preprint arXiv:1104.3250}, 2011.

\bibitem{bengio2013estimating}
Yoshua Bengio, Nicholas L{\'e}onard, and Aaron Courville.
\newblock Estimating or propagating gradients through stochastic neurons for
  conditional computation.
\newblock {\em arXiv preprint arXiv:1308.3432}, 2013.

\bibitem{neelakantan2015adding}
Arvind Neelakantan, Luke Vilnis, Quoc~V Le, Ilya Sutskever, Lukasz Kaiser,
  Karol Kurach, and James Martens.
\newblock Adding gradient noise improves learning for very deep networks.
\newblock {\em arXiv preprint arXiv:1511.06807}, 2015.

\bibitem{abadi2016tensorflow}
Mart{\'\i}n Abadi, Paul Barham, Jianmin Chen, Zhifeng Chen, Andy Davis, Jeffrey
  Dean, Matthieu Devin, Sanjay Ghemawat, Geoffrey Irving, Michael Isard, et~al.
\newblock Tensorflow: A system for large-scale machine learning.
\newblock In {\em 12th $\{$USENIX$\}$ symposium on operating systems design and
  implementation ($\{$OSDI$\}$ 16)}, pages 265--283, 2016.

\bibitem{bischl2017openml}
Bernd Bischl, Giuseppe Casalicchio, Matthias Feurer, Frank Hutter, Michel Lang,
  Rafael~G. Mantovani, Jan~N. van Rijn, and Joaquin Vanschoren.
\newblock Openml benchmarking suites, 2017.

\bibitem{scikit-learn}
F.~Pedregosa, G.~Varoquaux, A.~Gramfort, V.~Michel, B.~Thirion, O.~Grisel,
  M.~Blondel, P.~Prettenhofer, R.~Weiss, V.~Dubourg, J.~Vanderplas, A.~Passos,
  D.~Cournapeau, M.~Brucher, M.~Perrot, and E.~Duchesnay.
\newblock Scikit-learn: Machine learning in {P}ython.
\newblock {\em Journal of Machine Learning Research}, 12:2825--2830, 2011.

\bibitem{guyon2002gene}
Isabelle Guyon, Jason Weston, Stephen Barnhill, and Vladimir Vapnik.
\newblock Gene selection for cancer classification using support vector
  machines.
\newblock {\em Machine learning}, 46(1-3):389--422, 2002.

\bibitem{ghosh2019deep}
Kunal Ghosh, Annika Stuke, Milica Todorovi{\'c}, Peter~Bj{\o}rn J{\o}rgensen,
  Mikkel~N Schmidt, Aki Vehtari, and Patrick Rinke.
\newblock Deep learning spectroscopy: Neural networks for molecular excitation
  spectra.
\newblock {\em Advanced science}, 6(9):1801367, 2019.

\bibitem{cui2018modern}
Chenhao Cui and Tom Fearn.
\newblock Modern practical convolutional neural networks for multivariate
  regression: Applications to nir calibration.
\newblock {\em Chemometrics and Intelligent Laboratory Systems}, 182:9--20,
  2018.

\bibitem{zhang2019deepspectra}
Xiaolei Zhang, Tao Lin, Jinfan Xu, Xuan Luo, and Yibin Ying.
\newblock Deepspectra: An end-to-end deep learning approach for quantitative
  spectral analysis.
\newblock {\em Analytica chimica acta}, 1058:48--57, 2019.

\bibitem{sigurdsson2004detection}
Sigurdur Sigurdsson, Peter~Alshede Philipsen, Lars~Kai Hansen, Jan Larsen,
  Monika Gniadecka, and Hans-Christian Wulf.
\newblock Detection of skin cancer by classification of raman spectra.
\newblock {\em IEEE transactions on biomedical engineering}, 51(10):1784--1793,
  2004.

\bibitem{chen2004artificial}
Yi-ding Chen, Shu Zheng, Jie-kai Yu, and Xun Hu.
\newblock Artificial neural networks analysis of surface-enhanced laser
  desorption/ionization mass spectra of serum protein pattern distinguishes
  colorectal cancer from healthy population.
\newblock {\em Clinical Cancer Research}, 10(24):8380--8385, 2004.

\bibitem{charvat2020diffuse}
Jind{\v{r}}ich Charv{\'a}t, Ale{\v{s}} Proch{\'a}zka, Mat{\v{e}}j Fri{\v{c}}l,
  Old{\v{r}}ich Vy{\v{s}}ata, and Lucie Himmlov{\'a}.
\newblock Diffuse reflectance spectroscopy in dental caries detection and
  classification.
\newblock {\em Signal, Image and Video Processing}, pages 1--8, 2020.

\bibitem{selvaraju2017grad}
Ramprasaath~R Selvaraju, Michael Cogswell, Abhishek Das, Ramakrishna Vedantam,
  Devi Parikh, and Dhruv Batra.
\newblock Grad-cam: Visual explanations from deep networks via gradient-based
  localization.
\newblock In {\em Proceedings of the IEEE international conference on computer
  vision}, pages 618--626, 2017.

\bibitem{lime}
Marco~Tulio Ribeiro, Sameer Singh, and Carlos Guestrin.
\newblock "why should {I} trust you?": Explaining the predictions of any
  classifier.
\newblock In {\em Proceedings of the 22nd {ACM} {SIGKDD} International
  Conference on Knowledge Discovery and Data Mining, San Francisco, CA, USA,
  August 13-17, 2016}, pages 1135--1144, 2016.

\bibitem{lundberg2017unified}
Scott~M Lundberg and Su-In Lee.
\newblock A unified approach to interpreting model predictions.
\newblock In {\em Advances in neural information processing systems}, pages
  4765--4774, 2017.

\bibitem{guselnikova2019label}
O~Guselnikova, A~Trelin, A~Skvortsova, P~Ulbrich, P~Postnikov, A~Pershina,
  D~Sykora, V~Svorcik, and O~Lyutakov.
\newblock Label-free surface-enhanced raman spectroscopy with artificial neural
  network technique for recognition photoinduced dna damage.
\newblock {\em Biosensors and Bioelectronics}, 145:111718, 2019.

\bibitem{erzina2020precise}
M~Erzina, A~Trelin, O~Guselnikova, B~Dvorankova, K~Strnadova, A~Perminova,
  P~Ulbrich, D~Mares, V~Jerabek, R~Elashnikov, et~al.
\newblock Precise cancer detection via the combination of functionalized sers
  surfaces and convolutional neural network with independent inputs.
\newblock {\em Sensors and Actuators B: Chemical}, 308:127660, 2020.

\end{thebibliography}
\end{document}